\documentclass[preprint,12pt]{elsarticle}

\usepackage{filecontents}
\usepackage{balance}
\usepackage{epsfig}
\usepackage{cite}                       
\usepackage{amsfonts}
\usepackage{amssymb}
\usepackage{amsmath}
\usepackage{amsthm}
\usepackage{graphicx}                   
\usepackage{subfigure}                  
\usepackage{caption}
\usepackage{cases}                     
\usepackage{gensymb}                   
\usepackage{verbatim}                   
\usepackage{url}                      
\usepackage{algorithm}
\usepackage{algorithmicx}
\usepackage{color}
\usepackage{ifthen}
\newboolean{conclusions}
\setboolean{conclusions}{true}

\theoremstyle{plain}
\newtheorem{Theo}{Theorem}

\journal{Pattern Recognition}

\begin{document}
\begin{frontmatter}



\title{A Cross-Entropy-based Method to Perform Information-based Feature Selection}


\author[isti]{P.~Cassar\'a\corref{cor1}}
 \ead{pietro.cassara@isti.cnr.it}

 \author[waynaut]{A.~Rozza}
 \ead{alessandro.rozza@waynaut.com}

 \author[isti]{M.~Nanni}
 \ead{mirco.nanni@isti.cnr.it}

\cortext[cor1]{Corresponding author}

 \address[isti]{ISTI, CNR, Via Giuseppe Moruzzi 1, 56124, Pisa, Italy}
 \address[waynaut]{Waynaut, Research Team, Via Monte Nevoso 10, 20131, Milano, Italy}

\begin{abstract}
From a machine learning point of view, identifying a subset of relevant features from a real data set can be useful to improve the results achieved by classification methods and to reduce their time and space complexity. 
To achieve this goal, feature selection methods are usually employed. These approaches assume that the data contains redundant or irrelevant attributes that can be eliminated.
In this work, we propose a novel algorithm to manage the optimization problem that is at the foundation of the Mutual Information feature selection methods. Furthermore, our novel approach is able to estimate automatically the number of dimensions to retain.
The quality of our method is confirmed by the promising results achieved on standard real data sets.

\end{abstract}

\end{frontmatter}

\bibliographystyle{elsarticle-num}



\section{Introduction}
Since the 1960s, the rapid pace of technological advances allows to measure and record increasing amounts of data, producing real data sets comprising high-dimensional points.
Unfortunately, due to the ``curse of dimensionality''~\citep{Bellman1961}, high dimensional data are difficult to work with for several reasons.
First, a high number of features can increase the noise, and hence the error; secondly, it is difficult to collect an amount of observations large enough to obtain reliable classifiers since several classifiers do not properly deal with classification problems where the number of observations is lower than, or comparable to, the data dimensionality (see~\citep{Friedman1989,Chen2000,PatRec12}); 
finally, as the amount of available features increases, the space needed to store the data becomes high, while the speed of the employed algorithms could be too low. 

For all the above mentioned reasons, feature selection algorithms are often employed as the first pre-processing step to improve classification performances.
The aim of a feature selection technique is to identify a subset of relevant features to use in model construction. The main assumption when using a feature selection technique is that the data contains redundant or irrelevant attributes that can be omitted. 
Formally, given the input data matrix $\mathbf{X}$ composed by $n$ samples and $m$ features ($\mathbf{X} \in \Re^{n \times m}$), and the target classification vector
$\mathbf{y} \in \Re^{n}$, the feature selection problem is to find a $k$-dimensional subspace ($k \leq m$) by which we can characterize $\mathbf{y}$. 
Many interesting approaches have been proposed in literature~\citep{Chandrashekar2014} but most of them have a main drawback, they are not able to automatically estimate the value of $k$. 

In this work we present a novel algorithm to manage the optimization problem that is at the foundation of the Mutual Information (\emph{\textbf{MI}}, \citep{cover91, brown09}) feature selection methods. Moreover, the proposed approach is able to estimate automatically the value of $k$.
We develop an algorithm based on the Cross-Entropy approach, to filter out the set of features among all those available. We perform this operation through a stochastic approach, which allows us to select efficiently the minimal feature set $\mathbf{U} \subseteq \mathbf{X}$ that minimizes the conditional entropy between $\mathbf{U}$ and the class attribute $\mathbf{y}$. 
The quality of our method is confirmed by the promising results achieved on standard real data sets.

This paper is organized as follows: in Section~\ref{sec:related} the related works are presented; in Section~\ref{sec:theory} the theoretical background underlying our approach is summarized; in Section~\ref{sec:algorithm} our algorithm is described; in Section~\ref{sec:num_results} the experimental results are presented; finally, in Section~\ref{sec:conclusions} our conclusions are highlighted.

\section{Related Works}\label{sec:related}
From a machine learning point of view, if a classification algorithm uses irrelevant variables to generate the model, this model can be affected by poor generalization capabilities. 
To delete irrelevant features, a selection criterion is required. Once a feature selection criterion is selected, a procedure must be developed to find the subset of useful features. Directly evaluating all the subsets of features ($2^m$) for a given data is a NP-hard problem. For this reason, a suboptimal procedure must be employed to remove redundant data with tractable computations.

In~\citep{Kohavi1997}, the feature selection methods are classified into filter and wrapper methods. Filter methods compute the rank of the features employing different criterion and select the highly ranked features that are applied to the predictor. In wrapper methods the feature selection criterion is the performance of the predictor; precisely, the classifier is wrapped on a search algorithm which will find a subset which gives the highest predictor performance. 

Considering the filter feature selection approaches, one of the simplest criteria employed is the Pearson correlation coefficient~\citep{Guyon03} that is defined as follows:
\begin{equation}
R(j)=\frac{cov(\mathbf{x}_j,\mathbf{y})}{\sqrt{var(\mathbf{x}_j)*var(\mathbf{y})}}
\end{equation}%
where $\mathbf{x}_j$ is the $j$-th variable, $\mathbf{y}$ is the output (class labels), $cov(\cdot)$ is the covariance and $var(\cdot)$ the variance. 
It is important to notice that, this criteria can only detect linear dependencies between variable and target.

To overcame this limitation many approaches employ information theoretic ranking criteria to exploit the measure of dependency between variables.
The \textbf{\emph{MI}} between random variables is defined as follows: 
\begin{eqnarray}\label{MI_formula}
  \mathbf{I}(\mathbf{x};\mathbf{y})&=&\mathbf{H}(\mathbf{y})-\mathbf{H}(\mathbf{y}|\mathbf{x})=\mathbf{H}(\mathbf{x})-\mathbf{H}(\mathbf{x}|\mathbf{y})\nonumber\\
&=&\mathbf{H}(\mathbf{y})+\mathbf{H}(\mathbf{x})-\mathbf{H}(\mathbf{x},\mathbf{y})
\end{eqnarray}%
where $\mathbf{H}(\cdot)$ is the entropy function, and $\mathbf{x}$ or $\mathbf{y}$ or both can be matrixes or vectors.

One of the simplest \textbf{\emph{MI}}-based methods for feature selection is to find the \textbf{\emph{MI}}  between each feature and the output class labels and rank them based on this value. A threshold is set to select $k<m$ features. The results of this method can be poor (for further details see \citep{Forman2003}) since inter-feature \textbf{\emph{MI}} is not taken into account. Some approaches that improve the results of the aforementioned technique are proposed in~\citep{Battiti94}, \citep{Comon1994}, and \citep{Torkkola2003}.

Another example of this kind of algorithm is given in \citep{koller96}, where the authors model the classification procedure through the probability $Pr(\mathbf{y}|\mathbf{U})$ that is the probability to have the class $\mathbf{y}$ for the set of features $\mathbf{U}$. Under the assumption that there exists a subset of features $\mathbf{U_G}$ such that the probability $Pr(\mathbf{y}|\mathbf{U_G}) $ is close to $Pr(\mathbf{y}|\mathbf{U})$, the authors evaluate $\mathbf{U_G}$ minimizing the Kullback-Leibler's (\textbf{\emph{KL}}) divergence between the previous probabilities $D_K(Pr(\mathbf{y}|\mathbf{U_G}),Pr(\mathbf{y}|\mathbf{U}))$ over all possible sets of features.  
Exploiting the theory of Markov Blanket for the set of features, the authors provide an approximate methodology for the feature selection, which relies on the following considerations: for each feature $\mathbf{x}_j\in \mathbf{U_G}$, let $\mathbf{M}_j$ be the set of $k$ features $\mathbf{x}_i \in \mathbf{U_G}-\{\mathbf{x}_j\}$ for which the correlation between $\mathbf{x}_i$ and $\mathbf{x}_j$ has largest magnitude, the \textbf{\emph{KL}}-Divergence $D_K(Pr(\mathbf{y}|\mathbf{M}),Pr(\mathbf{y}|\mathbf{M}\cup \mathbf{x}_j))$ is evaluated for each $j$, the selected features are those one for which this quantity is minimal, then the set of selected feature is $ \mathbf{U_G} =\mathbf{U_G} - \{\mathbf{x}_j\}$.
This algorithm shows many weaknesses. First of all, due to the approximations adopted during the procedure (such as the Markov Blanket construction and \textbf{\emph{KL}}-Divergence evaluation), the algorithm provides a suboptimal solution. Moreover, the algorithm needs as input the size of the set of the selected features, and evaluates the correlation between all the possible pairs of features, therefore it cannot be efficient for domains containing hundreds or even thousands of features.

An improvement of the previous approach is based on the Mutual Information Maximization (\textbf{\emph{MIM}}) as proposed in \citep{brown09}. In this algorithm the features are ranked depending on the mutual information between each of them and the class attribute. \textbf{\emph{MIM}} supposes that all the features are independent and its criterion is to maximize the mutual information between feature $\mathbf{x}_j$ and class attribute $\mathbf{y}$. 

In\citep{Fleuret04} authors propose an extension of \textbf{\emph{MIM}} called Conditional Mutual Information Maximization (\textbf{\emph{CMIM}}). 
This approach iteratively picks features which maximize their mutual information with the class to predict, conditionally to the response of any feature already picked. This Conditional Mutual Information Maximization criterion does not select a feature similar to already picked ones, even if it is individually powerful, as it does not carry additional information about the class to predict. This criterion ensures a good tradeoff between independence and discrimination.

It is important to notice that, \textbf{\emph{MIM}} takes into account only the features that are relevant with the class variable (Maximal Relevance) but it does not consider any dependencies between these features. One desirable property is to have not features that are dependent since they produce redundant information. This is the main idea of the minimal-Redundancy-Maximal-Relevance criterion (\textbf{\emph{mRMR}}) \citep{peng05}. Precisely, this approach searches for the relevant features with the minimum redundancy. 

In \citep{Meyer06}, the authors introduce a new information theoretic criterion: the Double Input Symmetrical Relevance (\textbf{\emph{DISR}}). \textbf{\emph{DISR}} measures the symmetrical relevance of two features in all the possible combinations of a subset and selects among a finite number of subsets the most significant one. One advantage of \textbf{\emph{DISR}} is that a complementary feature of an already selected feature has higher probability to be chosen. 

In \citep{zheng11}, the authors measure the relationship between the candidate feature subsets $\mathbf{U}$ and the class attribute $\mathbf{y}$ exploiting the results on the Discrete Function Learning (\textbf{\emph{DFS}})  algorithm jointly to the high-dimensional mutual information evaluation. The method proposed by the author uses the entropy of the class attribute as the criterion to choose the appropriate number of features, instead of subjectively assigning the number of features a priori. Again, the authors provide the following interesting result: if the mutual information between a feature set $\mathbf{U}$ and the class attribute $\mathbf{y}$ is equal to the entropy of $\mathbf{y}$, then $\mathbf{U}$ is a Markov Blanket of $\mathbf{y}$.

\smallskip
Wrapper methods employ the predictor performance as the objective function to evaluate the variable subset. Since evaluating $2^m$ subsets is a NP-hard problem, suboptimal subsets are identified by employing heuristic search algorithms. The Branch and Bound method, proposed in \citep{Narendra1977}, uses tree structures to evaluate different subsets for the given feature selection number. Nevertheless, the search would grow exponentially for higher number of features. For these reasons, simplified algorithms such as sequential search or evolutionary algorithms (Genetic Algorithm \textbf{\emph{GA}}~\citep{Goldberg1989} or Particle Swarm Optimization \textbf{\emph{PSO}}~\citep{Kennedy95}) which yield local optimum results are employed. The \textbf{\emph{PSO}} is a population-based search technique and is motivated by the social behavior of organisms. It is based on swarm intelligence and well suited for combinatorial optimization problems in which the optimization surface possesses many local optimal solutions. The underlying phenomenon of \textbf{\emph{PSO}} is that knowledge is optimized by social interaction where the thinking is not only personal but also social. The particles in \textbf{\emph{PSO}} resemble to the chromosomes in \textbf{\emph{GA}}. However, \textbf{\emph{PSO}} is usually easier to implement than the \textbf{\emph{GA}} as there are neither crossover nor mutation operators in the \textbf{\emph{PSO}} and the movement from one solution set to another is achieved through the velocity functions.These approaches can produce good results and are computationally feasible but are strongly related to the selected predictor.

Most of the filter methods stated above perform the feature selection basing their research on the maximization/minimization of the amount of information carried by the selected features about the class attribute. In this case Information Theory provides a solid mathematical framework to state the problem, that can be implemented through an algorithm based on many different optimization approaches, such as robust, genetic, or Bayesian-based. The main advantages offered by the algorithms based on Information Theory are that the features with redundant information are filtered out, and the performances of these algorithms do not degrade for high-dimensional sets of features. Moreover, the solid mathematical framework allows a rigorous performance evaluation of the developed techniques.
Most of the the algorithms discussed above overcome many of the problems related to the feature selection: they have a sound theoretical foundation, they are effective in eliminating both irrelevant and redundant features, they are tolerant to inconsistencies in the training data, and finally, they are filter methods which do not suffer the high dimensionality of the space of features. 
The main shortcomings lie in: candidate feature is evaluated pairwise (step by step) with respect to every individual feature in the selected feature subset; the number of features $k$ needs to be specified a-priori. The motivation for the first shortcoming is that the feature $\mathbf{x}_j$ is good only if it carries information about the class attribute $\mathbf{y}$ and such information has not been caught by any of the features already picked. However, it is unknown whether the existing features as a vector have captured the information carried by $\mathbf{x}_j$ or not. Due to the second shortcoming, the performances of most of the state-of-the-art algorithms can be affected by an imprecise selection of the value of $k$. 

Our algorithm overcomes all the weakness shown by these algorithms, in fact, all the estimated variables are evaluated through optimized algorithms and are not approximated; the candidate feature is not evaluated pairwise with respect to every individual feature in the selected feature subset; finally, the algorithm estimates autonomously the dimension of the set of selected features.

\section{Theoretical Background}
\label{sec:theory}

In this section we introduce the theoretical background related to the feature selection approaches based on the information theoretic ranking. These concepts are exploited to explain the optimization problem related to the feature selection addressed in this paper. In Section~\ref{sec:algorithm} we provide a statistical interpretation of the optimization problem, and then, we propose a Cross-Entropy  based method to find a solution for that. 

We start recalling the meaning of Mutual Information (\textbf{\emph{MI}}). Mutual information can be used to measure the amount of information obtained about a random variable $\mathbf{A}$, through another one $\mathbf{B}$. We remember that \textbf{\emph{MI}} is a symmetric non-negative function. Furthermore, the \textbf{\emph{MI}} is strictly related to the entropy $\mathbf{H}(\mathbf{A})$ of a random variable, which defines the amount of information held in $\mathbf{A}$. Note that, the entropy of $\mathbf{A}$ grows if its outcomes have low probability to arise, otherwise the entropy decreases. Hence we can assert that the entropy measures the diversity of $\mathbf{A}$ in terms of uncertainty of its outcomes. The \textbf{\emph{MI}} between random variables is defined as in the equation (\ref{MI_formula}). 

In our case study, we are interested to the \textbf{\emph{MI}} between the subset of features $\mathbf{U}=\{\mathbf{x}_1 \cdots \mathbf{x}_k \; | \; k\leq m\} \subseteq \mathbf{X}  $, and the class attribute $\mathbf{y}$ : 

\begin{equation} 
  \label{MI_3}
  \mathbf{I}(\mathbf{U};\mathbf{y})=\mathbf{H}(\mathbf{y})-\mathbf{H}(\mathbf{y}|\mathbf{U})
\end{equation}

where $\mathbf{H}(\mathbf{y}|\mathbf{U})$ is the amount of information needed to describe $\mathbf{y}$ conditioned by the information held in  $\mathbf{U}$ about $\mathbf{y}$. Hence, this quantity represents the dependence between $\mathbf{U}$ and $\mathbf{y}$, precisely, the greater the information that can be obtained about $\mathbf{y}$ through $\mathbf{U}$, the lowest the information needed about $\mathbf{y}$ once that $\mathbf{U}$ has been known. This means that the features in $\mathbf{U}$ can fully determine the values of $\mathbf{y}$. In this case the features into the set $\mathbf{U}$ are called essential attributes (\textbf{\emph{EA}}s). This means that if the features in $\mathbf{U}$ are \textbf{\emph{EA}}s, then the equation (\ref{MI_3}) gets is maximum.
The considerations discussed above have been summarized in the theorem that follows next, which has been proven in  \citep{eliece77, cover91}.
 
 \begin{Theo} \label{theorem dependence of y}
   If the \textbf{\emph{MI}} between $\mathbf{U}$ and $\mathbf{y}$ is equal to the entropy of $\mathbf{y}$, then $\mathbf{y}$ is function of $\mathbf{U}$.
 \end{Theo}

The quantities in equation (\ref{MI_3}) can be evaluated trough  a set of $n$ samples of both class attribute and subset of selected features, then, 
$\mathbf{y},\mathbf{x}_j \in \Re^{n}$ for $j=1\cdots k$, and $\mathbf{U} \in \Re^{n \times k}$. A powerful tool for the entropy and mutual information functions  evaluation can be found in \citep{MItb}. 

Now, we explain the information maximization criteria used for the information-based ranking. As stated above, the  \textbf{\emph{MI}} measures the amount of information obtained about a random variable $\mathbf{y}$, through $\mathbf{U}$. As proven in the papers \citep{cover91, eliece77} the  $\mathbf{I}(\mathbf{U};\mathbf{y})$ can be written as:

\begin{gather}
   \label{MI property 3}
   \mathbf{I}(\mathbf{x}_1\cdots \mathbf{x}_k;\mathbf{y})=\displaystyle \sum_{j=1}^{k} \mathbf{I}(\mathbf{x}_j;\mathbf{y}|\mathbf{x}_{j-1}\cdots \mathbf{x}_1)\\
   \nonumber
    \mathbf{U}_{k} = \{ \mathbf{x}_1 \cdots \mathbf{x}_k\}\\
   \label{joined MI}
         \mathbf{I}(\mathbf{U}_{k};\mathbf{y})= \mathbf{I}(\mathbf{U}_{k-1};\mathbf{y})+\mathbf{I}(\mathbf{x}_k;\mathbf{y}|\mathbf{U}_{k-1}).
   \end{gather}

\begin{figure}[ht!]
\begin{center}
  \includegraphics[width=9cm]{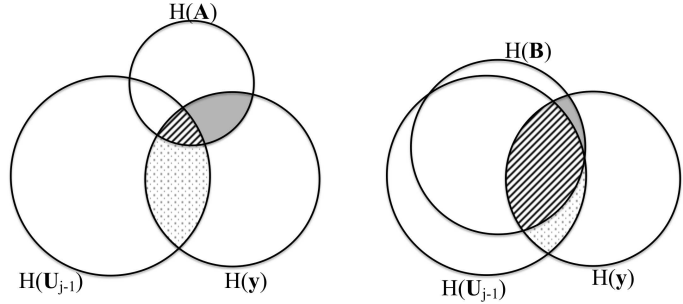}
  \caption{Example of the relationship between Mutual Information and Entropy} \label{figure_example} 
 \end{center}
 \end{figure}

Equations (\ref{MI property 3}-\ref{joined MI}) allow to select iteratively the features in $\mathbf{U}$, so that the \textbf{\emph{MI}} can be maximized. Precisely, in the first step we select the j'-th feature $\mathbf{x}_{j'} \;|\; 1 \leq j' \leq m$ that shares the largest \textbf{\emph{MI}} with the class attribute $\mathbf{y}$, searching within the native set. During the other iterations, the feature $\mathbf{x}_j$  is selected if it adds the maximum information to the already selected features $\{\mathbf{x}_1 \cdots \mathbf{x_{j-1}}\} \;|\;j=1 \cdots k \leq m$, by searching into the native set minus the features already selected. An intuitive proof of equations (\ref{MI property 3}-\ref{joined MI}) is given through the example in figure \ref{figure_example}. In this example the circles are the entropies of the variables, and the gray regions are the \textbf{\emph{MI}} between the feature that can be selected and the class attribute $\mathbf{y}$, given the already selected features $\mathbf{U}_{j-1}$. Hence, the feature to be selected is $A$ in this example. In figure \ref{figure_example},  the sum of dashed and  gray areas is the information shared between $\mathbf{y}$  and the new feature $A$ or $B$, respectively. Instead, the shared information between $\mathbf{U}_{j-1}$ and  $\mathbf{y}$ is the sum of the dashed and dotted areas. Then, if we want to evaluate the information  about $\mathbf{y}$ carried by the new feature given the set  $\mathbf{U}_{j-1}$, we need to take into account only the gray area, which is greater in the case of $A$ than in the case of $B$. This means that $\mathbf{U}_{j-1}$ contains most of the information carried by $B$ about $\mathbf{y}$, then $B$ is redundant and can be eliminated. In fact, the dashed area is greatest in the case of $B$. On the opposite, the feature $A$ can be saved because carries new information about $\mathbf{y}$ not yet contained in $\mathbf{U}_{j-1}$. 

 Hence, selecting iteratively the features of $\mathbf{U}$ satisfying the equality $\mathbf{I}(\mathbf{U}; \mathbf{y}) = \mathbf{H}(\mathbf{y})$, $\mathbf{U}$ can be taken as complete  subset for the prediction of $\mathbf{y}$ (see theorem \ref{theorem dependence of y}).  Note that the equalities discussed above are obtained at the limit, hence, the problem of finding optimal feature subsets  is equivalent to find the subset $\mathbf{U}$, over the native set, subject to  $\mathbf{I}(\mathbf{U}; \mathbf{y}) \rightarrow \mathbf{H}(\mathbf{y} )$, or equivalently subject to $  \mathbf{H}(\mathbf{y}| \mathbf{U}) \rightarrow 0$.
 
 The results discussed above are very important, because they allow us to design our optimized algorithm for the selection of the optimal features. In the next section, we provide the mathematical framework, and then  the new algorithm for the search of the subset of features that minimizes $  \mathbf{H}(\mathbf{y}| \mathbf{U})$, avoiding the research of the solution on a set of $\binom{m}{k}$ elements, given that these are all possible set with cardinality $k$ for a native set with cardinality $m$. 
 
\section{The Algorithm Formalization}
\label{sec:algorithm}
In this section we present a novel and effective algorithm to find the solution for the feature selection problem based on the information theoretic ranking, presented in the previous section. As stated above, this search can be performed through our optimized procedure based on the mathematical framework described in the following. 

The approach adopted in this work is based on the stochastic research of the solution. The main idea of this approach is to associate to each feature $\mathbf{x}_i$ ($i=1,\cdots,m$) a binary variable $z_i \in \{0,1\}$ that can take value $1$ with probability $p_i$. This is called Associated Stochastic Problem \textbf{\emph{ASP}}  \citep{rubinstein04}. Our algorithm finds which variables  $z_i, \: i=1,\cdots,m$ must have $p_i \rightarrow 1$, so that the objective function $\mathcal{O}(\mathbf{U}(\mathbf{z}))$ given by $\mathbf{I}(\mathbf{U}; \mathbf{y})$ is maximum. This means to find the optimal density distribution  of the binary vector $\mathbf{z}$ that has the $i-th$ entry to $1$ if the $i-th$ feature must be selected to maximize  $\mathbf{I}(\mathbf{U}; \mathbf{y})$. Then, we can perform the search of the solution on a probability space through the \textbf{\emph{ASP}}, and no longer on the initial space given by all the possible combinations of indexes of the features. Then, the combinatorial problem based on equations (\ref{MI property 3}-\ref{joined MI})  has become a convex problem (for more details see Appendix A) through the \textbf{\emph{ASP}}.

Hence,  we start defining the \textbf{\emph{ASP}} for our initial maximization problem. Let  $\mathbf{z}=[z_1 \cdots z_m ]$ be a binary vector of cardinality $m$, where $z_i=1$ if the $i-th$ feature belongs to $\mathbf{U}$ and $0$ otherwise. Hence, we can rewrite the subset of the selected features as function of $\mathbf{z}$. In this case, the objective function $\mathcal{O}(\mathbf{U}(\mathbf{z}))$ for our  is again a function of $\mathbf{z}$. Our goal is to forecast which entries of $\mathbf{z}$ must be set to $1$ to maximize $\mathcal{O}(\mathbf{U}(\mathbf{z}))$ subject to $\sum_{i=1}^m z_i=k \leq m$. The last step to define our \textbf{\emph{ASP}} is to assume that $z_1 \cdots z_m$ are independent Bernoulli random variables where the probabilities to get the value $1$ are $p_1 \cdots p_m$, respectively. Then the evaluation problem of the distribution of entries equal to $1$ in $\mathbf{z}$, for which  
the objective function is maximum $\displaystyle \max_{\mathbf{z}}\{\mathcal{O}(\mathbf{U}(\mathbf{z}))\}=\gamma$, can be formulated as in the following: to find the distribution $g(\mathbf{z},v)$ of the $1$ values in $\mathbf{z}$, under a given parameter $v$, which maximizes the probability for the objective function to be greater or equal to $\gamma$, \textit{i.e.} $P_v(\mathcal{O}(\mathbf{U}(\mathbf{z})) \geq \gamma)$.

Note that $P_v(\mathcal{O}(\mathbf{U}(\mathbf{z}))\geq \gamma)$ can be estimated through the following equation given the probability distribution function $g(\mathbf{z},v)$: 

\begin{equation} 
  \label{probability estimation}
         P_v(\mathcal{O}(\mathbf{U}(\mathbf{z}))\geq \gamma)=\sum_{\mathbf{z}}\mathcal{I}_{\{\mathcal{O}(\mathbf{U}(\mathbf{z}))\geq \gamma\}} \:g(\mathbf{z},v)
 \end{equation}%
where  $\mathcal{I}_{\{\cdot\}} $ is the indicator function of the event $\mathcal{O}(\mathbf{U}(\mathbf{z}))\geq \gamma$. The indicator function is equal to $1$ for all the possible configurations of $\mathbf{z}$ that  verify the event $\mathcal{O}(\mathbf{U}(\mathbf{z}))\geq \gamma$, and $0$ otherwise.

A well known approach to estimate the probability in the equation (\ref{probability estimation}) is to use the Likelihood Ratio (\textbf{\emph{LR}}) estimator with reference parameter $v$. From the theory of \textbf{\emph{LR}} estimator \citep{rubinstein04} the best value $v^*$, such that the samples $\mathbf{z}$ drawn from $g(\mathbf{z},v^*)$ provide the maximum value $\gamma$ for our objective function $\mathcal{O}(\mathbf{U}(\mathbf{z}))$, can be calculated through the following equation:

\begin{equation} 
  \label{parameter estimation}
         v^*=\displaystyle arg\max_{v} \frac{1}{S} \sum_{j=1}^{S}\mathcal{I}_{\{\mathcal{O}(\mathbf{U}(\mathbf{z_j}))\geq \gamma\}} \:\mbox{ln}(g(\mathbf{z_j},v))
 \end{equation}%
where $\mathbf{Z}=\{\mathbf{z}_1, \cdots, \mathbf{z}_S\}$ is a set of possible  samples drawn by the distribution $g(\mathbf{z},v)$ and $\mathbf{z}_j=[z_{1j} \cdots z_{mj} ]$.

As stated above $\mathbf{z}_j=[z_{1j} \cdots z_{mj} ]$ is a vector of independent Bernoulli random variables where $z_{ij}$ takes value equal to $1$ with probability $p_{i}$ and $0$ with probability $1-p_{i}$. Then for the assumptions stated above, the probability distribution $g(\mathbf{z},v)$ of the ones in $\mathbf{z}$ can be rewritten with parameter $v=\mathbf{p}=[p_{1} \cdots p_{m}]$ as in the following:

\begin{equation} 
  \label{pdf}
                           g(\mathbf{z},\mathbf{p})=\displaystyle \prod_{i=1}^m p_i^{z_{i}}(1-p_i)^{(1-z_{i})} \: ; \: z_{i} \in \{0,1\}
 \end{equation}
 
Given that for equation (\ref{parameter estimation}) it holds the convexity property, its solution can be found in a closed form through derivation w.r.t. $p_i$ and imposing the derivative to be equal to $0$.

\begin{gather} 
  \nonumber
 \frac{\partial{}}{\partial{p_i}}\frac{1}{S} \sum_{j=1}^{S}\mathcal{I}_{\{\mathcal{O}(\mathbf{U}(\mathbf{z_j}))\geq \gamma\}} \:\mbox{ln}(g(\mathbf{z_j},v))=0 \rightarrow \\
\nonumber
\frac{1}{(1-p_i)p_i} \sum_{j=1}^{S}\mathcal{I}_{\{\mathcal{O}(\mathbf{U}(\mathbf{z_j}))\geq \gamma\}}(z_{ij}-p_i)=0 \rightarrow \\
\label{closed form solution}
p_i=\frac{\sum_{j=1}^{S}\mathcal{I}_{\{\mathcal{O}(\mathbf{U}(\mathbf{z_j}))\geq \gamma\}}z_{ij}}{\sum_{j=1}^{S}\mathcal{I}_{\{\mathcal{O}(\mathbf{U}(\mathbf{z_j}))\geq \gamma\}}}\:\;\: i=1 \cdots m;
\end{gather}

The equation (\ref{closed form solution}) allows us to evaluate the vector $\mathbf{p}$, where its entries closer to $1$ correspond to the entries of $\mathbf{z}$ that with high probability take the value $1$.

The derivation of equations \ref{parameter estimation}-\ref{closed form solution} is shown in the \textbf{Appendix A} where the optimality of the probability distribution function $g(\mathbf{z},\mathbf{p})$  is also proven.

Summarizing, we found the density distribution of the  entries in $\mathbf{z}$ that take the value $1$, these entities provide the set of the indexes of the features that maximize the objective function $\mathcal{O}(\mathbf{U}(\mathbf{z}))$ that  is the mutual information  $\mathbf{I}(\mathbf{U}; \mathbf{y})$ .

The solution for the probability $\mathbf{p}$ can be calculated through an iterative algorithm as showed in the following (see Algorithm \ref{algo: main_CE}).

\begin{algorithm} [h!]
\caption{CE-based Algorithm for Entries Probability Calculation with Parameters Statically Set}
\label{algo: main_CE}
\begin{algorithmic}
\State \emph{1.  Define the initial distribution $g(\mathbf{z},\mathbf{p}^0)$ where $\mathbf{p}^0$ is chosen}\\
\hspace{7pt}\emph{uniform,  the step of the algorithm is set to 1 ($t=1$);}\\
\hspace{7pt}\emph{Set $\forall i\leq 0: \gamma_{i}=\infty$;}
\medskip
\State \emph{While$\left(|\gamma_t-\gamma_{t-d}|\geq \epsilon\right)$}
\medskip
\State \hspace{14pt} \emph{2.}\\
\hspace{18pt} \emph{a  Generate the set of samples $\mathbf{z}_1 \cdots \mathbf{z}_S \sim g(\mathbf{z},\mathbf{p}^{t-1})$,}\\
\hspace{20pt}\emph{b The objective functions $ \mathcal{O}(\mathbf{U}(\mathbf{z_j}))\; j=1 \cdots S$ are}\\ 
\hspace{29pt}\emph{evaluated as well as their $(1-\rho)$-quantile,}\\ 
\hspace{20pt}\emph{c $\gamma_t$  is set to the $ (1-\rho)$-quantile  of the objective }\\
\hspace{29pt}\emph{functions $ \mathcal{O}(\mathbf{U}(\mathbf{z_j}))$}
\medskip
\State \hspace{14pt} \emph{3. The vector $\mathbf{p}^t=[p_1^t \cdots p_m^t]$ is evaluated through the\\ \hspace{24pt}equation (\ref{closed form solution}) by the }\emph{samples $\{\mathbf{z}_j\} \;i = 1\cdots S$, \\ \hspace{24pt}for which $ \mathcal{O}(\mathbf{U}(\mathbf{z}_j))$ fall to their $(1-\rho)$-quantile};
\medskip
\State \hspace{14pt} \emph{4. The step t is incremented by one;} 
\State \emph{End}
\medskip
\State \emph{5. After t' steps, when $|\gamma_{t'}-\gamma_{t'-d}|\leq \epsilon$ we get the vector} \\
\hspace{6pt} \emph{ of probability distribution $\mathbf{p}^{t'} $ and then $g(\mathbf{z},\mathbf{p}^{t'})$  is}\\
\hspace{6pt} \emph{evaluated trough equation (\ref{pdf});}
\medskip
\State \emph{6. The vector $\mathbf{z}$ of the indexes of the selected feature is evaluated}\\ 
 \[  z_{i}=\left \{  \begin{array}{cc}
                                                 1 &  \\
                                                 0 &  
                         \end{array}  \sim g(\mathbf{z},\mathbf{p}^{t'}) \;i=1 \cdots m \right. ;\]

\end{algorithmic}
\end{algorithm}

The Algorithm \ref{algo: main_CE} estimates iteratively the value $\mathbf{p}$, and hence it helps to evaluate  the vector $\mathbf{z}$ of the indexes of the selected features that maximize the function $\mathcal{O}(\mathbf{U}(\mathbf{z}))$. The algorithm finds the optimal value of the probability vector $\mathbf{p}^{t'}$ after $t'$ steps, starting from the uniform distribution $\mathbf{p}^0=\mathit{uniform}$  at step $t=0$. The distribution probability $\mathbf{p}^t$  at each step $t$ is refreshed trough the samples $\mathbf{z}_1 \cdots \mathbf{z}_S$ drawn from the distribution $g(\mathbf{z},\mathbf{p}^{t-1})$ and the equation (\ref{closed form solution}). Precisely, the events $\mathcal{O}(\mathbf{U}(\mathbf{z_j}))\geq \gamma_t$ are verified over the samples $\{\mathbf{z}_1 \cdots \mathbf{z}_S\}$, and then, in the equation (\ref{closed form solution}) are used only those samples belonging to the $(1-\rho)$-quantile that verifies the inequality $\mathcal{O}(\mathbf{U}(\mathbf{z_j}))\geq \gamma_t$. This means that the maximum $\gamma_t$ of the objective function is refreshed with its  $(1-\rho)$-quantile step by step. The stopping criteria for the algorithm is stated as follows: if the max value of the objective function does not change by at least $\epsilon$ after $d$ steps, the optimal distribution probability $\mathbf{p}^{t'}$ is found. Assuming that the algorithm finds $\mathbf{p}^{t'}$ after $t'$ steps, then $k\leq m$ entries of $\mathbf{p}^{t'}$ will be very close to $1$ and $m-k$ entries will be very close to $0$.

The parameters $S$ and $\rho$ can be statically set as in the Algorithm \ref{algo: main_CE}, or, they can be dynamically computed at each step. In the first case the tuning of these parameters can be found in \citep{rubinstein04}. In the last case (which is ours), the parameter $S$ can be evaluated generating multiple sequences  of samples $\{\mathbf{z}_1 \cdots \mathbf{z}_{S_{min}}\} \cdots \{\mathbf{z}_1 \cdots \mathbf{z}_{S_{max}}\}$  at each step $t$, by ranging their length between $S_{min}$ and $S_{max}$. The value $S_{min}$ is set to the cardinality of the unknown parameters ($m$ in our case) and the value $S_{max}$ is set to the empirical value $20\cdot S_{min}$. For each sequence of length $S_h\in [S_{min}\cdots S_{max}]$, the event $\mathcal{O}(\mathbf{U}(\mathbf{z_j}))\geq \gamma_t$ is verified, then the length $S_h$ adopted at the step $t$ is the largest  which generates the $(1-\rho)$-quantile . 
At the same time, the quantile parameter at the step $t$ is evaluated through the empirical formula $\rho=0.05 \cdot m/S_h$. Finally, the threshold parameter $\epsilon$ is set to the empirical value of $5\%$ (for further details about the tuning parameters see \citep{rubinstein04}, chapter 5). 


Through $\mathbf{p}$ we can evaluate the distribution $g(\mathbf{z},\mathbf{p})$ of the ones in $\mathbf{z}$, i.e.  the best set of features for our maximization problem. Moreover, through $g(\mathbf{z},\mathbf{p})$ we are able to study the behavior of the error of the classification model, for different size of the set of the selected features (for more details see \citep{rubinstein04}).   

\section{Experimental Results}
\label{sec:num_results}
The performance evaluation of our algorithm has been performed on
six data sets from UCI Machine Learning Repository \citep{uci}. 
All experiments have been performed on an iMAC endowed with a 2,4 GHz Intel Core 2 Duo, and 4 GB of SDRAM at 667 MHz. All the involved algorithms have been implemented in MATLAB.

The chosen data have integer and real values for the class attributes, and they can have single or multi-label class attributes.
Considering the feature values, they can take integer and real values, and the cardinality of features set ranges between medium to wide size. 

\begin{table}[htb]
{\scriptsize
\begin{center}

\begin{tabular}{|c|c|c|c|} \hline
  & \textbf{Advertisements} & \textbf{Blog Feedback} & \textbf{Cancer Breast}  \\ \hline
 \textbf{num. samples}   & 3279 & 52397& 569  \\ \hline
 \textbf{num features}    & 1558 & 280 & 32 \\ \hline
 \textbf{feature type }     & Binary/Real& Real/Integer & Real  \\ \hline
  \textbf{attribute type }   & Binary & Integer & Binary   \\ \hline
  & \textbf{Connect-4} & \textbf{Forest Fires} & \textbf{Gesture Phase} \\ \hline
 \textbf{num. samples }  & 67557 & 517 & 1444   \\ \hline
 \textbf{num features }   & 42 & 12 & 32  \\ \hline
 \textbf{feature type}      & Integer & Real& Real  \\ \hline
 \textbf{attribute type}    & Integer & Real & Integer \\ \hline

\end{tabular}
\end{center}}
\caption{Data Sets}
\label{tab:data_sets}
\end{table}

Precisely, the Advertisements dataset represents a set of possible advertisements on Internet pages.  The features encode the geometry of the image, if available, the phrases occurring in the URL, the image's URL, the alt text, the anchor text, and words occurring near the anchor text. The task is to predict whether an image is an advertisement `1' or not `0'.

The Blog Feedback data set contains the data originated from blog posts. The raw HTML-documents of the blog posts were crawled and processed. The prediction task associated with the data is to identify the number of comments in the upcoming 24 hours. 
Precisely, 
the features represent the average, standard deviation, min, max and median of the parameters, and others parameters which characterize the number of comments in the time, such as number of comments in the first 24 hours after the publication of the blog post, and the length of the blog post. The class attribute is the number of comments in the following 24 hours.

The Cancer Breast data are computed from a digitized image of a fine needle aspirate of a breast mass. They describe characteristics of the cell nuclei present in the image. The features are real values and represent the image characteristics of the nuclei, such as radius, texture, smoothness. The class attribute is a binary value that is the diagnosis malignant `1' or not `0'.

The Connect-4 data set contains all legal 8-ply positions in the game of \emph{connect-4} game, in which neither player has won yet, and in which the next move is not forced. The features are integer values and represent the positions taken by one of the two players on a matrix with six rows and seven columns. Precisely, if the player 1 takes the position $a_i$ of the matrix $a_i=1$, if the position is taken by the player 2 $a_i=2$, if it is blank $a_i=3$. The class attribute is an integer value that is the result of the game: `1' or `0' if one of the two players won, or `2' if it was a draw.

The Forest Fires data set contains information about all the parameters that can be involved in the forest fires. The features are real values and represent the spatial coordinates of the forest, the date, the parameters of the most famous forest fire danger rating systems, and the weather parameters. The class attribute is a real value that is the burned area of the forest. 

The data set Gesture Phase contains the data about the temporal segmentation of gestures, performed by gesture researchers in
order to pre-process videos for further analysis. The data set is composed by the data that come from  videos recorded using Microsoft Kinect sensor. The collected features represent an image of each frame identified by a timestamp, the text file containing positions coordinates $x$, $y$, $z$ of six articulation points: left hand, right hand, left wrist, right wrist, head and spine, with each line in the file corresponding to a frame and identified by a timestamp. The class attribute is an integer value that is the gesture phase: Rest `1', Preparation `2', Stroke `3', Hold `4', Retraction `5'.
Table \ref{tab:data_sets} summarizes the data sets used during the performance evaluation.

 \begin{table}[htb]
{\scriptsize
\tabcolsep=1mm
\begin{center}
\begin{tabular}{|c|c|c|c|c|} \hline
       &                                       & \textbf{Advertisements} & \textbf{Blog Feedback} & \textbf{Cancer Breast}  \\ \hline
\textbf{Our}                              & $\textbf{MCE}_{NB}$           & \textbf{0.1055}                              & \textbf{0.3837}                              & \textbf{0.0371}  \\ \cline{2-5}
			                             & $\textbf{MCE}_{KNN}$           & \textbf{0.0037}                             & \textbf{0.0018}                              & \textbf{0.01 } \\ \cline{2-5}

                                                & \textbf{$\Delta I_r$}         & \textbf{0.002}                            & 0.2965                              & 25.0797\\ \cline{2-5}
                                                & \textbf{$\Delta t$}     & 176.29                          & 178.12                              & 162.54\\ \hline
\textbf{DISR}                            & $\textbf{MCE}_{NB}$   &0.21                          & 0.5071                             &0.0726  \\ \cline{2-5}
										& $\textbf{MCE}_{KNN}$           & 0.1459               & 0.3888              & \textbf{0.01}  \\ \cline{2-5}
                                                & \textbf{$\Delta I_r$}          & 0.1038                             &0.3414                                & 24.7977\\ \cline{2-5}
                                                & \textbf{$\Delta t$}     & \textbf{167.46}                         & 184.62                           &184.43\\ \hline
\textbf{CMIM}                          & $\textbf{MCE}_{NB}$           & 0.1169                         & 0.4364                              & 0.0856\\ \cline{2-5}
 									   & $\textbf{MCE}_{KNN}$   & 0.0065 & 0.22                  & \textbf{0.01}  \\ \cline{2-5}
                                                & \textbf{$\Delta I_r$}          & 0.0096                             & \textbf{0.2945 }                               & \textbf{23.1734} \\ \cline{2-5}
                                                & \textbf{$\Delta t$}     & 174.17                         & \textbf{164.44}                            &172.63\\ \hline
\textbf{mRMR}                         & $\textbf{MCE}_{NB}$           & //                           & 0.4979                              & 0.0705  \\ \cline{2-5}
									  & $\textbf{MCE}_{KNN}$   & 0.0103 & 0.1568                  & \textbf{0.01}  \\ \cline{2-5}
                                       
                                                & \textbf{$\Delta I_r$}          & 0.0181                              & 0.3725                                & 23.9602 \\ \cline{2-5}
                                                & \textbf{$\Delta t$}    & 184.43                        & 184.76                              &180.51\\ \hline
& \textbf{ Cardinality} & 39 &93  &20  \\ \hline      
      
 &                                      & \textbf{Connect-4}        & \textbf{Forest Fires}       & \textbf{Gesture Phase} \\ \hline
 \textbf{Our}                            & $\textbf{MCE}_{NB}$           &\textbf{0.3904}                             &\textbf{0.0775}                             & \textbf{0.2888}  \\ \cline{2-5}
										& $\textbf{MCE}_{KNN}$           & \textbf{0.00067}                             & \textbf{0.052 }                             & \textbf{0.01}  \\ \cline{2-5}
                                              & \textbf{$\Delta I_r$}        & \textbf{0.0011}                              &\textbf{ 0.0115}                                & 0.0244 \\ \cline{2-5}
                                               & \textbf{$\Delta t$}     & 178.27                          & \textbf{172.65}                             &161.95\\ \hline
\textbf{DISR}                          & $\textbf{MCE}_{NB}$           & 0.4361                              & 0.0789                             & 0.2915 \\ \cline{2-5}
										& $\textbf{MCE}_{KNN}$           & 0.1457               & 0.068              & \textbf{0.01}  \\ \cline{2-5}
                                               & \textbf{$\Delta I_r$}        & 0.3648                              &0.0532                               & \textbf{0.0241} \\ \cline{2-5}
                                                & \textbf{$\Delta t$}     & \textbf{164.05}                      & 180.31                              &\textbf{161.39}\\ \hline
\textbf{CMIM}                         & $\textbf{MCE}_{NB}$           & 0.4267                              & 0.0789                             & 0.5591\\ \cline{2-5}
									  & $\textbf{MCE}_{KNN}$   & 0.0228 & 0.063                  & \textbf{0.01}  \\ \cline{2-5}
                                       
                                              & \textbf{$\Delta I_r$}        & 0.0396                              & 0.0532                               & \textbf{0.0241}\\ \cline{2-5}
                                               & \textbf{$\Delta t$}     & 171.04                         & 184.49                             &181.73\\ \hline
\textbf{mRMR}                       & $\textbf{MCE}_{NB}$           & 0.4406                              & 0.0899                             & 0.3099  \\ \cline{2-5}
									& $\textbf{MCE}_{KNN}$   		& 0.1691 & 0.0797                  &\textbf{ 0.01 } \\ \cline{2-5}
                                       
                                             & \textbf{$\Delta I_r$}        & 0.4591                              & 0.0455                       & 0.0306  \\ \cline{2-5}
                                              & \textbf{$\Delta t$}     & 183.39                         & 176.89                            &183.85\\ \hline
                                             & \textbf{ Cardinality} & 32 &5 &29  \\ \hline   
\end{tabular}
\end{center}
}

\caption{Performance Comparison.  `$// $' means that a pooled estimation of the covariance or diagonal covariance matrix can not be evaluated through the naive Bayes classifiers using only the features selected by that method.}
\label{tab:perf_comp}
\end{table}

To assess the performance of our algorithm we have generated independent training and validation sets from the aforementioned data sets.
The training and test sets are independent for the Blog Feedback data set, instead, for the others data set the training and test sets have been evaluated through the partition $90\%$ and $10\%$, respectively. 
As stated in Section \ref{sec:algorithm}, parameters $S$ and $\rho$ are dynamically computed at each step of the algorithm. Instead, entries  $p_1^0 \cdots p_m^0 $ of the initial probability vector have been set all to $0.5$, and the exit condition for the algorithm has been set to $|\gamma_t-\gamma_{t-d}| \geq 0.05$ with $d=5$, as suggested in \citep{rubinstein04}. 

The performances of our algorithm have been compared with those achieved by the following Information-based feature selection algorithms: \textbf{\emph{DISR}}, \textbf{\emph{CMIM}}, and \textbf{\emph{mRMR}}. For the performance evaluation the parameters taken into account are: the misclassification error (\textbf{\emph{MCE}}), the relative difference of information ($\Delta I_r$), and the execution time ($\Delta t$) in seconds. The \textbf{\emph{MCE}} is the number of misclassified observations divided by the number of observations on the test set as a function of the number of features; the $\Delta I_r$  is the ratio between the difference $\mathbf{I}(\mathbf{U};\mathbf{y})-\mathbf{H}(\mathbf{y})$ and the Mutual Information $\mathbf{I}(\mathbf{U};\mathbf{y})$; and the $\Delta t$ is merely the time spent to get the set of selected features. Through the \textbf{\emph{MCE}} we can study the accuracy of the given classification model, using the selected features. Instead, through the $\Delta I_r$ we can study how well the class attribute can be represented as function of the set of selected features. Precisely, the class attribute is function of the selected features for $\Delta I_r$ close to zero. 

For the evaluation of the \textbf{\emph{MCE}} we have taken into account a classification model that fits the multivariate normal density to each group, with a pooled estimate of covariance or with a diagonal covariance matrix estimate through Naive Bayes classifiers. Moreover, we have also employed the KNN classifier (with $k=3$) thus to use a different predictor to fully evaluate the proposed method.

Table \ref{tab:perf_comp} summarizes the performance results in terms of \textbf{\emph{MCE}} (both employing Naive Bayes, $\textbf{\emph{MCE}}_{NB}$, and KNN, $\textbf{\emph{MCE}}_{KNN}$), $\Delta I_r$, and $\Delta t$, where the bold entries are the best results obtained. These results show how our algorithm, in all the data set analyzed, is able to find the optimal set of features which minimizes $\Delta I_r$. Note that our method in some cases provides one order of accuracy more than other techniques. Moreover, in the cases where this does not happen, it is due to the fact that our algorithm minimizes also the \textbf{\emph{MCE}} for the optimal set of selected features. The rows labelled as \textbf{Cardinality} shows the cardinality of the feature set automatically selected through our approach\footnote{This \textbf{cardinality} is used to evaluate the set of features selected by the other algorithms taken into account in this work.}. Considering $\Delta t$, our approach always shows a comparable time cost\footnote{Notice that, the implementation of our approach is not fully optimize whilst for the other algorithms we have used optimized toolbox implementations.}.  

\begin{figure}[!h]
\begin{center}
  \includegraphics[width=13 cm]{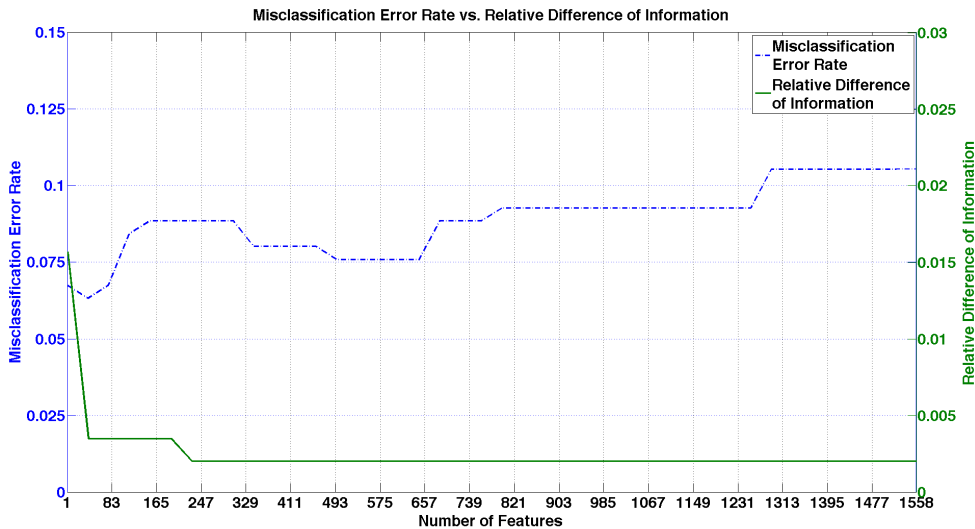}
  \caption{The \textbf{\emph{MCE}} (using Naive Bayes) and $\Delta I_r$ functions computed employing our approach on Advertisements data set.} \label{figure_comparison_ad} 
 \end{center}
 \end{figure}

\begin{figure}[!h]
\begin{center}
  \includegraphics[width=13cm]{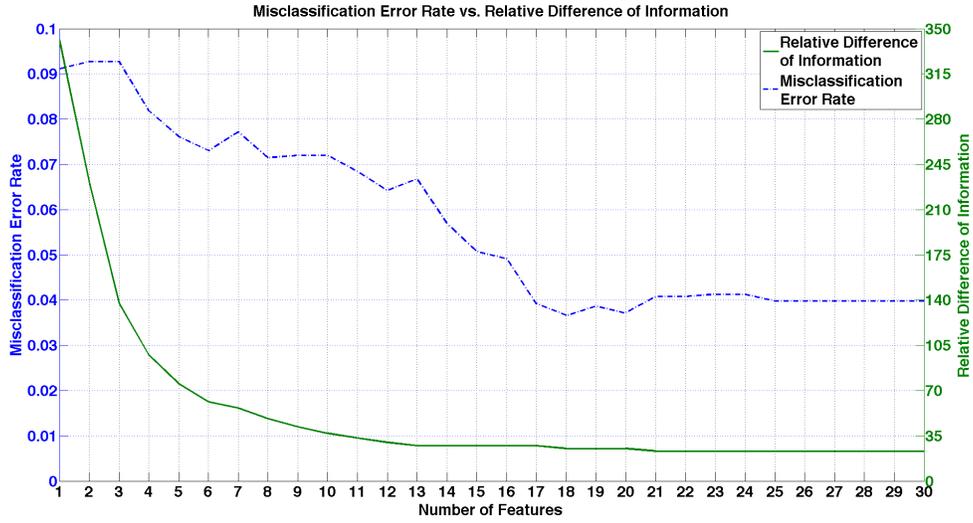}
  \caption{The \textbf{\emph{MCE}} (using Naive Bayes) and $\Delta I_r$ functions computed employing our approach on Cancer Breast data set.} \label{figure_comparison_cancer} 
 \end{center}
 \end{figure}

\begin{figure}[!h]
\begin{center}
  \includegraphics[width=13 cm]{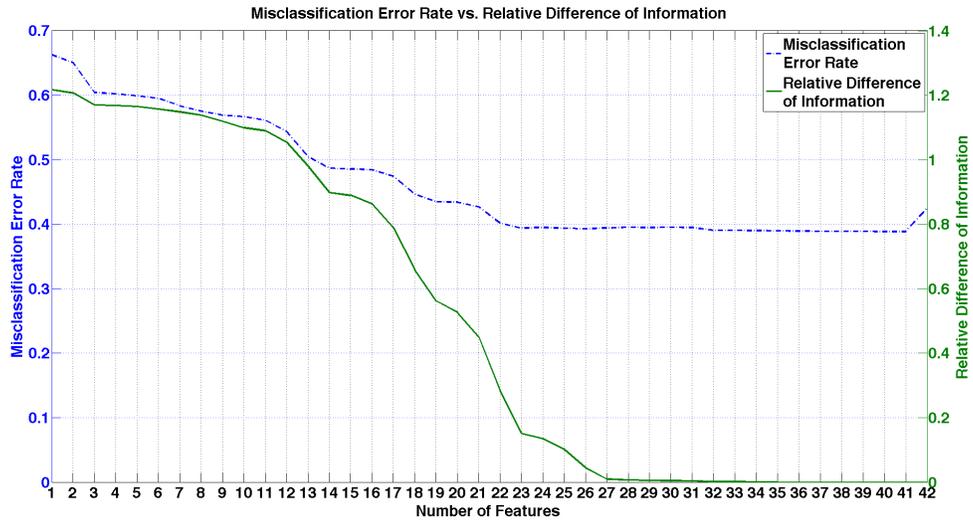}
  \caption{The \textbf{\emph{MCE}} (using Naive Bayes) and $\Delta I_r$ functions computed employing our approach on Connect-4 data set.} \label{figure_comparison_connect} 
 \end{center}
 \end{figure}

\begin{figure}[!h]
\begin{center}
  \includegraphics[width=13 cm]{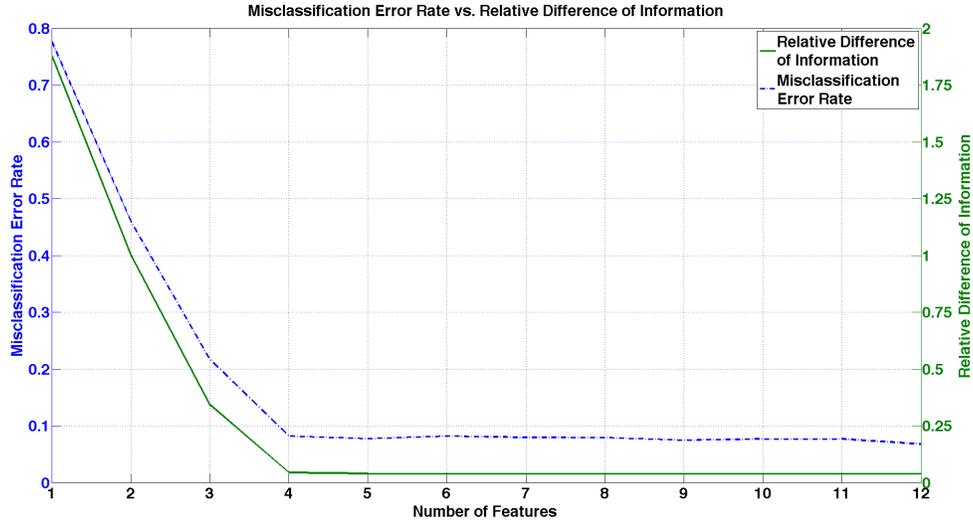}
  \caption{The \textbf{\emph{MCE}} (using Naive Bayes) and $\Delta I_r$ functions computed employing our approach on Forest Fires data set.} \label{figure_comparison_fires} 
 \end{center}
 \end{figure}

\begin{figure}[!h]
\begin{center}
  \includegraphics[width=13 cm]{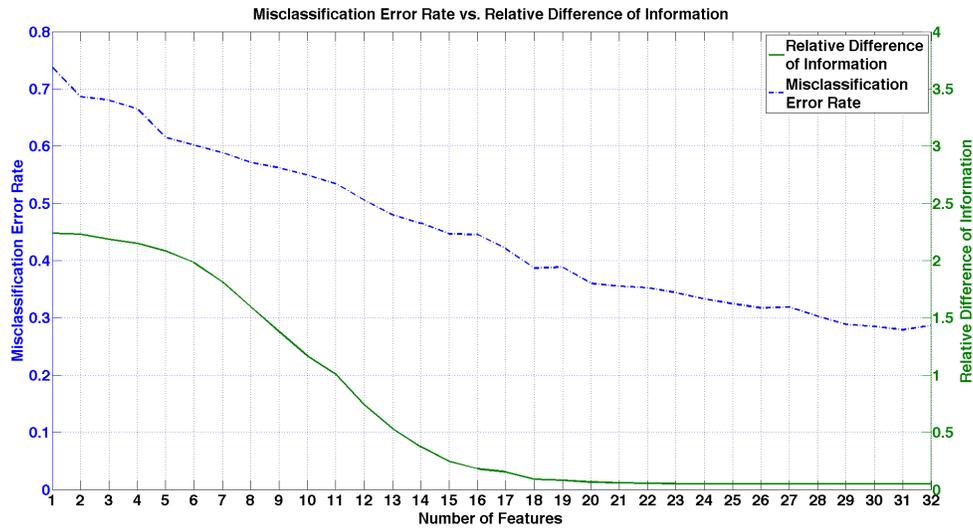}
  \caption{The \textbf{\emph{MCE}} (using Naive Bayes) and $\Delta I_r$ functions computed employing our approach on Gesture Phase data set.} \label{figure_comparison_gesture} 
 \end{center}
 \end{figure}

The first clear advantage given by our algorithm is the ability to find the best minimum set of features that maximize $\mathbf{I}(\mathbf{U};\mathbf{y})$. Moreover, our algorithm finds the set of features that minimizes at the same time the \textbf{\emph{MCE}} and the $\Delta I_r$ as shown in the figures \ref{figure_comparison_ad}-\ref{figure_comparison_gesture}. Each figure shows the \textbf{\emph{MCE}} and the $\Delta I_r$ as functions of different values of retained features. 
These figures show that the chosen set of features as well as their cardinality are the best for the addressed problem, indeed, for the chosen cardinality the \textbf{\emph{MCE}} and the $\Delta I_R$ have the global minimum. 

\begin{figure}[!h]
\begin{center}
  \includegraphics[width=13 cm]{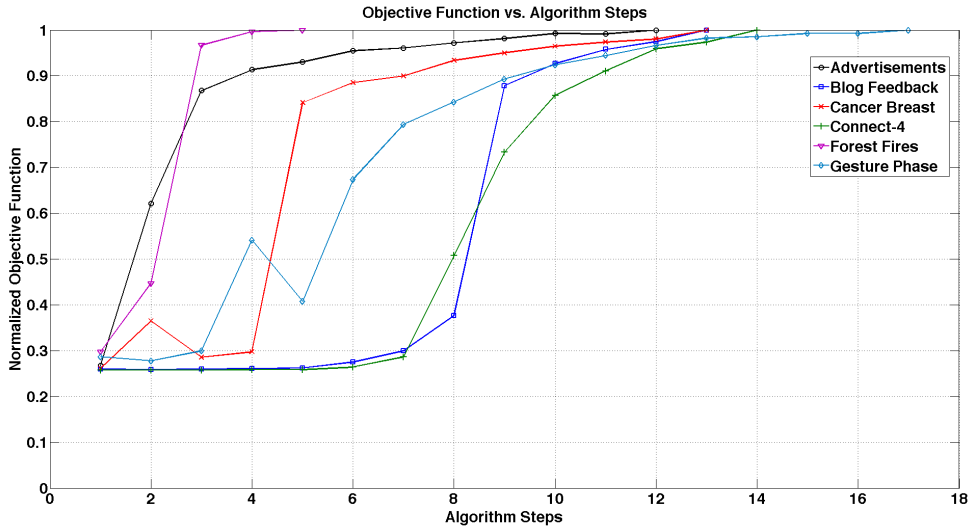}
  \caption{Number of Steps Needed for the Maximization of the Objective Function} \label{figure_convergence} 
 \end{center}
 \end{figure}

\begin{table}[htb]
\scriptsize
\tabcolsep=1mm
\begin{center}
\begin{tabular}{|c|c|c|c|c|} \hline
                                                &                                         & \textbf{Advertisements}   & \textbf{Blog Feedback}    & \textbf{Cancer Breast}  \\ \hline
\textbf{Our}                              & \textbf{MCE}                    &\textbf{ 0.1055}               & \textbf{0.3837}                 & \textbf{0.0371}  \\ \cline{2-5}
                                                & \textbf{$\Delta I_r$}         & \textbf{0.002}                   & \textbf{0.2965}                 & 25.0797\\ \cline{2-5}
                                                & \textbf{$\Delta t$}            & \textbf{176.29}                 & \textbf{178.12}                 & \textbf{162.54}\\ \hline
\textbf{PSO}                            & \textbf{MCE}                    & 0.1441                             & 0.4851                              & 0.0523  \\ \cline{2-5}
                                                & \textbf{$\Delta I_r$}         & 0.0084                              & 0.3345                              & \textbf{23.6741} \\ \cline{2-5}
                                                & \textbf{$\Delta t$}            &302.67                              & 289.32                               &198.78\\ \hline
& \textbf{ Cardinality CE/PSO}                                           & \textbf{39}/42                  &\textbf{93}/87                      &\textbf{20}/18  \\ \hline      
      
                                               &                                              &\textbf{Connect-4}      & \textbf{Forest Fires}     & \textbf{Gesture Phase} \\ \hline
 \textbf{Our}                            & \textbf{MCE}                         &\textbf{0.3904}         &\textbf{0.0775}              & 0.2888  \\ \cline{2-5}
                                              & \textbf{$\Delta I_r$}               & \textbf{0.0011}         & \textbf{0.0115}            &\textbf{ 0.0244} \\ \cline{2-5}
                                               & \textbf{$\Delta t$}                 & 178.27                    & \textbf{172.65 }            &\textbf{161.95}\\ \hline
\textbf{PSO}                        & \textbf{MCE}                            & 0.4345                     & 0.0878                        & \textbf{0.266}  \\ \cline{2-5}
                                             & \textbf{$\Delta I_r$}                & 0.0525                     & 0.0492                        & 0.0287  \\ \cline{2-5}
                                              & \textbf{$\Delta t$}                  & \textbf{174.32}       & 201.02                         &196.21\\ \hline
                                             & \textbf{ Cardinality CE/PSO}   & \textbf{32}/27         &\textbf{5}/ \textbf{5}      &\textbf{29}/35  \\ \hline   
\end{tabular}
\end{center}
\caption{Performance Comparison between CE-based algorithm and PSO using Naive Bayes as classifier.}
\label{tab:perf_comp_2}
\end{table}

We have also compared our algorithm  with the Particle Swarm Optimization method (PSO, \citep{Eberhart97}), which is one of the most representative heuristic methods, since the heuristic approaches can be suitable to address this kind of problem. The comparison results are shown in table \ref{tab:perf_comp_2}, where the bold entries are again the best results obtained.

The results show that our algorithm has better performance especially in terms of $\Delta t$. In fact, the PSO algorithm selects the subset of features $\mathbf{U}$ by a random proportional rule, and even if the algorithm analyzes the research history (see \citep{Eberhart97}), its convergence is slower with respect to our algorithm. The main advantage given by our approach is that we refresh the joint occurrence probability of the features step by step, which provides faster and more accurate convergence compared with the proportional rule.

We have investigated also the ability of our algorithm to converge rapidly towards the solution, for all the datasets taken into account in this paper. Precisely, figure \ref{figure_convergence} shows the number of steps required by our algorithm to maximize the normalized  objective function, i.e. the mutual information. For all the datasets taken into account, the figure shows that in $10$ steps in the average, our algorithm is able to find the optimal subset of features. Moreover, the algorithm converges toward the solution stably, with power law. 


\section{Conclusions}
\label{sec:conclusions}
In this work, we have presented a novel algorithm to manage the optimization problem that is at the foundation of the Mutual Information feature selection methods. A peculiarity of this novel approach consists in the ability to estimate automatically the dimensions to retain.
The proposed methodology is based on a Cross-Entropy approach that filters out the set of features among all those available. 
The numerical results obtained on standard real data sets have confirmed that our method is promising.

\balance
\bibliography{biblio.bib}




\appendix
\section{Derivation of Equations 12 and 13} \label{appA}

Consider the following general maximization problem. Let $\mathbb{U}$ be a finite set
of states, and let $\mathcal{O}$ be a real-valued performance function on $\mathbb{U}$. The issue is to find the maximum of $\mathcal{O}$  over  $\mathbb{U}$, and the corresponding state(s) at which this maximum is attained. Let us denote the maximum by $\gamma^*$ thus to obtain:

\begin{equation} 
  \label{eq_1_app}
           \mathcal{O}(\mathbf{U}^*)=\gamma^* = \displaystyle \max_{\mathbf{U} \in \mathbb{U}}  \mathcal{O}(\mathbf{U})
 \end{equation}

This equation represents the associate estimation problem related to the starting optimization problem.

To find the solution through the Cross-Entropy Theory, we define a collection of indicator functions $\mathcal{I}_{\{\mathcal{O}(\mathbf{U})\} }$ on $\mathbb{U}$ for various thresholds $\gamma \in \Re$. Next, let $\{g(\cdot,\mathbf{v}), \mathbf{v} \in \mathbb{V}\}$ be a family of probability distribution functions (pdf) on $\mathbb{U}$, parameterized by a real-valued parameter (vector) $\mathbf{v}$. Given $\mathbf{v} \in \mathbb{V}$ we associate with equation \ref{eq_1_app} the problem of estimating the following quantity:

\begin{gather} 
      \nonumber   P_{\mathbf{v}}(\mathcal{O}(\mathbf{U})\geq \gamma)=\\ \sum_{\mathbf{u}}\mathcal{I}_{\{\mathcal{O}(\mathbf{u})\geq \gamma\}} \:g(\mathbf{u},\mathbf{v})=\mathbb{E}_{\mathbf{v}}\:\mathcal{I}_{\{\mathcal{O}(\mathbf{U})\geq \gamma\}}  \label{eq_2_app}
 \end{gather}%
where $ P_{\mathbf{v}}$ is the probability measure under which the random state $\mathbf{U} $ has pdf $g(·;\mathbf{v})$, and $\mathbb{E}_{\mathbf{v}}$ denotes the corresponding expectation operator. The estimation problem in equation \ref{eq_2_app} is called the associated stochastic problem (ASP). To indicate how (\ref{eq_2_app}) is associated with (\ref{eq_1_app}), suppose for example that $\gamma$ is equal to $\gamma^*$  and that $g(·;\mathbf{v})$ is the uniform density on $\mathbb{U}$. Note that, the way to evaluate $P_{\mathbf{v}}(\mathcal{O}(\mathbf{U})\geq \gamma^*)$ is through its occurrence rate, hence  $P_{\mathbf{v}}(\mathcal{O}(\mathbf{U})\geq \gamma^*)= g(\mathbf{U}^∗;\mathbf{v}) = 1/|\mathbb{U}|$, where $|\mathbb{U}|$ denotes the number of elements in $\mathbb{U}$. Thus, for $\gamma= \gamma^*$ a natural way to estimate $ P_{\mathbf{v}^*}(\mathcal{O}(\mathbf{U})\geq \gamma^*)$ would be to use an estimator based on the Kulback-Leibler distance and with optimal reference parameter $\mathbf{v}^*$, as shown in the following.

The expectation operator $\mathbb{E}_{\mathbf{v}}$ can be written by the Lebesgue-integral as: 

\begin{equation} \label{eq_3_app}
\mathbb{E}_{\mathbf{v}}\:\mathcal{I}_{\{\mathcal{O}(\mathbf{U})\geq \gamma\}}=\int_{\mathbb{U}} \:\mathcal{I}_{\{\mathcal{O}(\mathbf{U})\geq \gamma\}} g(\mathbf{u},\mathbf{v}) d\mathbf{u}
\end{equation}

A straightforward way to estimate this integral is to use crude Monte-Carlo simulation drawing a random sample $\mathbf{U}_1, . . . , \mathbf{U}_N$ from $g(· ;\mathbf{v})$, but this approach is inefficient because a large simulation effort is required in order to estimate accurately the expectation, i.e., with small relative error or a narrow confidence interval. An alternative, based on importance sampling, takes a random sample $\mathbf{U}_1, . . . , \mathbf{U}_N$ from an importance sampling density $g(· ;\mathbf{w})$ on $\mathbb{U}$, and evaluates the expectation using this new estimator:

\begin{gather} 
\nonumber \mathbb{E}_{\mathbf{v}}\:\mathcal{I}_{\{\mathcal{O}(\mathbf{U})\geq \gamma\}}=\int_{\mathbb{U}} \:\mathcal{I}_{\{\mathcal{O}(\mathbf{U})\geq \gamma\}} \frac{g(\mathbf{u},\mathbf{v})}{g(\mathbf{u} ;\mathbf{w})} g(\mathbf{u} ;\mathbf{w}) d\mathbf{u}=\\
\mathbb{E}_{\mathbf{w}}\:\mathcal{I}_{\{\mathcal{O}(\mathbf{U})\geq \gamma\}}\frac{g(\mathbf{u},\mathbf{v})}{g(\mathbf{u},\mathbf{w})}. \label{eq_4_app}
\end{gather}

It is well known (see \citep{rubinstein04}) that one of the best ways to estimate the expectation is to use the change of measure with density:

\begin{equation}  \label{eq_5_app}
g(\mathbf{u} ;\mathbf{w})=\frac{\mathcal{I}_{\{\mathcal{O}(\mathbf{U})\geq \gamma\}}g(\mathbf{u} ;\mathbf{v})}{ \mathbb{E}_{\mathbf{v}}\mathcal{I}_{\{\mathcal{O}(\mathbf{U})\geq \gamma\}}}
\end{equation}

Note that $g(\mathbf{u} ;\mathbf{w})$ (called \textit{optimal Importance Sampling distribution}) depends on the unknown value $ \mathbb{E}_{\mathbf{v}}\mathcal{I}_{\{\mathcal{O}(\mathbf{U})\geq \gamma\}}$, so this probability distribution function is hard to evaluate.

For this kind of problem the Cross-Entropy method provides an iterative method to evaluate the optimal density $g(\mathbf{u} ;\mathbf{w})$. In brief, the method evaluates $g(\mathbf{u} ;\mathbf{w'})$, minimizing the Kulback-Leibler distance between the optimal densities $g(\mathbf{u} ;\mathbf{w})$ and $g(\mathbf{u} ;\mathbf{w'})$:

\begin{gather}    
  \nonumber D_k(g(\mathbf{u} ;\mathbf{w'}),g(\mathbf{u} ;\mathbf{w}))=\int_{\mathbb{U}}  \ln{\frac{g(\mathbf{u},\mathbf{w'})}{g(\mathbf{u} ;\mathbf{w})}} g(\mathbf{u} ;\mathbf{w' }) d\mathbf{u}=\\
\int_{\mathbb{U}} g(\mathbf{u} ;\mathbf{w'}) \ln{g(\mathbf{u},\mathbf{w'})} d\mathbf{u}-\int_{\mathbb{U}} g(\mathbf{u} ;\mathbf{w'}) \ln{g(\mathbf{u},\mathbf{w})} d\mathbf{u} \label{eq_6_app}
\end{gather}

The minimization of the equation (\ref{eq_6_app}) is equivalent to the maximization of the following quantity:

\begin{equation}    \label{eq_7_app}
\int_{\mathbb{U}} g(\mathbf{u} ;\mathbf{w'}) \ln{g(\mathbf{u},\mathbf{w})} d\mathbf{u}\end{equation}

Hence, substituting $g(\mathbf{u} ;\mathbf{w'}) $ with the optimal solution in equation \ref{eq_5_app} we obtain the maximization program:

\begin{gather}
  \nonumber \displaystyle \max_{\mathbf{w}} \int_{\mathbb{U}} \frac{\mathcal{I}_{\{\mathcal{O}(\mathbf{U})\geq \gamma\}}g(\mathbf{u} ;\mathbf{v})}{ \mathbb{E}_{\mathbf{v}}\mathcal{I}_{\{\mathcal{O}(\mathbf{U})\geq \gamma\}}} \ln{g(\mathbf{u} ;\mathbf{w})} d\mathbf{u}=\\
  \displaystyle \max_{\mathbf{w}} \mathbb{E}_{\mathbf{v}}\:\mathcal{I}_{\{\mathcal{O}(\mathbf{U})\geq \gamma\}} \ln{g(\mathbf{u} ;\mathbf{w})}
  \end{gather}%
that is the problem addressed in the equation (\ref{parameter estimation}), with solution in equation (\ref{closed form solution}). Note that the maximization problem involves a convex function, so the optimal solution is unique if it exists.

\end{document}